\pdfoutput=1

\documentclass[11pt]{article}

\usepackage[final]{acl}

\usepackage{times}
\usepackage{latexsym}
\usepackage{graphicx}
\usepackage{xcolor}
\usepackage{amssymb}

\usepackage{colortbl}
\usepackage{pgf}
\usepackage{stfloats}

\newcommand{\MinNumber}{-0.01} 
\newcommand{\MaxNumber}{0.8}

\newcommand{\ApplyGradient}[1]{%
    \pgfmathsetmacro{\PercentColor}{max(min(100.0*(#1-\MinNumber)/(\MaxNumber-\MinNumber),100.0),0.00)} 
    \edef\temp{\noexpand\cellcolor{green!\PercentColor!white}}\temp 
    \temp #1 
}

\newcommand{\gradecolor}[2]{%
    \ApplyGradient{#1} #2 
}

\usepackage[T1]{fontenc}

\usepackage[utf8]{inputenc}

\usepackage{microtype}

\usepackage{inconsolata}

\title{Your Transformer is Secretly Linear}

\author{
Anton Razzhigaev\textsuperscript{1,2},  
Matvey Mikhalchuk\textsuperscript{1,5}, 
Elizaveta Goncharova\textsuperscript{1,4}, \\
\bf Nikolai Gerasimenko\textsuperscript{3,5},
\bf Ivan Oseledets\textsuperscript{1,2}, 
\bf Denis Dimitrov\textsuperscript{1,3}, and
Andrey Kuznetsov\textsuperscript{1,3}\\
\textsuperscript{1}AIRI, 
\textsuperscript{2}Skoltech,
\textsuperscript{3}SberAI,
\textsuperscript{4}HSE University,\\
\textsuperscript{5}Lomonosov Moscow State University\\
\href{mailto:razzhigaev@skol.tech}{razzhigaev@skol.tech} \\ %
}

\begin{document}
\maketitle
\begin{abstract}

This paper reveals a novel linear characteristic exclusive to transformer decoders, including models such as GPT, LLaMA, OPT, BLOOM and others. We analyze embedding transformations between sequential layers, uncovering a near-perfect linear relationship (Procrustes similarity score of 0.99). However, linearity decreases when the residual component is removed due to a consistently low  output norm of the transformer layer. Our experiments show that removing or linearly approximating some of the most linear blocks of transformers does not affect significantly the loss or model performance. Moreover, in our pretraining experiments on smaller models we introduce a cosine-similarity-based regularization, aimed at reducing layer linearity. This regularization improves performance metrics on benchmarks like Tiny Stories and SuperGLUE and as well successfully decreases the linearity of the models. This study challenges the existing understanding of transformer architectures, suggesting that their operation may be more linear than previously assumed.\footnote{\url{https://github.com/AIRI-Institute/LLM-Microscope}}

\end{abstract}

\section{Introduction}

Transformers have revolutionized the field of natural language processing, offering unprecedented advances in a wide range of applications \cite{islam2023comprehensive}. However, despite their widespread adoption and success, the complex work of these models remains an area of active research \cite{lin2021survey}. One aspect that has received less attention is the inherent linearity of intermediate embedding transformations within these architectures. In this study, we embark on an in-depth analysis of the linearity properties of transformers, specifically focusing on decoders, and explore its implications during the pretraining and fine-tuning phases.

Our investigation reveals a surprising discovery: the embedding transformations between sequential layers in transformer decoders exhibit almost linear properties. This observation is quantified using Procrustes similarity analysis, demonstrating a near-perfect linearity score of 0.99. Such a discovery not only challenges the traditional understanding of transformer architectures but also opens new opportunities for model optimization and efficiency.

Based on this insight, we introduce several new contributions to the field:
\begin{itemize}
    \item Extensive analysis of the linearity properties of transformer decoders and its dynamics at the pretraining and fine-tuning stages.
    \item The development of new algorithms for depth pruning of transformer decoders, allowing to remove the most linear layers without a significant loss in performance.
    \item A novel distillation technique that involves pruning, replacing certain layers with linear approximations, and then distilling layer-wise embeddings to preserve model performance.
    \item Introducing a new regularization approach for pretraining based on the cosine similarity, designed to decrease the layer linearity. This method not only enhances the performance of transformer models on benchmark datasets such as SuperGLUE and TinyStories \citet{eldan2023tinystories}, but also improves the expressiveness of embeddings, as evidenced by linear probing tasks.
\end{itemize}

 With our findings, we are paving the way for more computationally efficient transformer architectures without sacrificing their effectiveness, thereby addressing one of the critical challenges in deploying these models.

\begin{figure*}
    \centering

    \includegraphics[width=\linewidth]{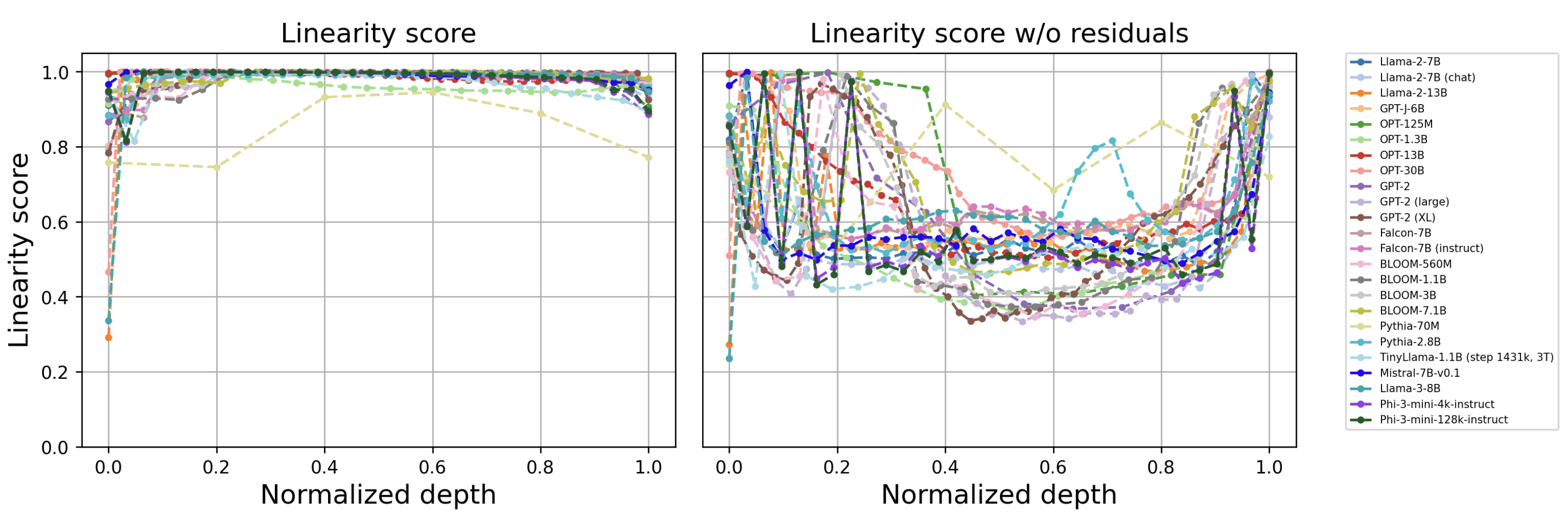}
    \caption{Linearity profiles for different open source models. Normalized depth is the layer index divided by the total depth.}
    \label{fig:linearity_profiles}
\end{figure*}

\section{Related Work}

Research on evaluating and leveraging sparsity for model pruning has become one of the most significant topics within the machine learning community. \citet{molchanov2016pruning} explored the sparsity of convolutional neural networks through backpropagation and fine-tuning, laying the groundwork for understanding the potential applications of sparsity in resource-efficient inference. The verification approach utilized in a more recent DejaVu \cite{borse2023dejavu} paper is based on Molchanov's research.


 Previous work \cite{kurtic2023sparse} has addressed the challenges associated with naive sparse fine-tuning in the context of LLMs. Issues such as training instability, poor recovery, and overfitting have prompted an exploration for alternative approaches. The study introduced SquareHead distillation, a method that consistently addresses the challenges in naive sparse fine-tuning, demonstrating accurate recovery even at high sparsity levels.

In a more recent study WANDA \cite{sun2023wanda}, the authors present a technique for pruning LLMs to high degrees of sparsity without modifying the remaining weights. Unlike SparseGPT \cite{frantar2023sparsegpt}, WANDA seamlessly implements pruning in a single forward pass, leveraging feature norm statistics for efficient pruning. This method achieves noticeable sparsity without the need for a sophisticated iterative weight update procedure, differentiating itself from other pruning techniques.

Contextual sparsity introduced by \citet{borse2023dejavu} involves sparsifying MLP and attention blocks in LLMs to reduce generation latency. The study identifies essential attention heads and MLP neurons for computation, maintaining performance across in-context learning and language modeling tasks. 

Recent work by \citet{ashkboos2024slicegpt} shows that LLMs can be sparsified post hoc. Their approach introduces a scheme to replace each weight matrix with a smaller dense matrix, thereby reducing the dimensionality of the networks. Their results show that models of different sizes can be reduced with varying degrees of success. For example, LLAMA-2 70B and OPT 66B can maintain 99\% zero-shot accuracy while reducing 25\% of the parameters reduced while performing LLM evaluation tasks. In contrast, the smaller Phi-2 is more sensitive to pruning, experiencing a 10\% drop compared to its dense version.

The inner structure of transformer models has captured significant attention among researchers \cite{nostalgebraist2020, xu-etal-2021-probing, belrose2023eliciting, din2023jump}. Primarily, in ``logit lens'' \cite{nostalgebraist2020} and subsequently in \cite{belrose2023eliciting}, the authors have focused on analyzing how hidden representations evolve across different layers of transformer architecture, aiming to elucidate their impact on final model outputs. Complementing these findings, the Anthropic team's research into small transformer-based models \cite{anthropic2021} uncovers a profound linear structure inherent in this architecture. Their work demonstrates the effectiveness of decomposing operations into individual sum components and multiplying chains of matrices, thus highlighting the linear complexity within these sophisticated neural architectures.

\paragraph{Structure-based pruning}

Topological features that analyze the structure of inner embeddings in transformer-based models are also useful in LLM pruning and distillation. 
Previous research examined the intrinsic dimensionality of neural networks to evaluate their capacity and effectiveness in the fine-tuning process \cite{ansuini2019intrinsic,aghajanyan2020intrinsic,razzhigaev2023shape}. Decoder-based models are shown to achieve a high level of anisotropy, especially in their middle layers, and have low intrinsic dimensionality \cite{razzhigaev2023shape}. Recent popular approaches include low-rank approximation, which replaces or adjusts the weight matrix with the product of two matrices with a smaller inner dimension. This approach typically requires a fine-tuning procedure that adjusts the matrix representations. For example, LoRA \cite{hu2021lora} was inspired by the previous work \cite{aghajanyan2020intrinsic} showing that neural networks can be successively trained in lower-dimensional subspaces. The research also shows that there it is not necessary to update millions of parameters on small fine-tuning datasets. Our results are on par with the results of this research, showing that via fine-tuning, the linearization of models grows steadily.

The Bonsai model \cite{dery2024everybody} tends to prune the LLMs relying only on the inference step, while they achieve performance comparable to half-sized semistructured sparsity with WANDA 2:4 and outperforms the LLM-Pruner \cite{ma2023llmpruner} and LoRAPrune \cite{zhang2023loraprune} on 4 out of 6 evaluation settings in the experiments conducted.

In this paper, we investigate several techniques for pruning LLMs, leveraging the linearity of the decoder-based layers. Our techniques offer efficient yet lightweight methods, maintaining high model performance on the evaluated benchmarks.

\section{Analysis of Pretrained Architectures}

\begin{figure}[h!]
    \centering
    \includegraphics[width=\linewidth]{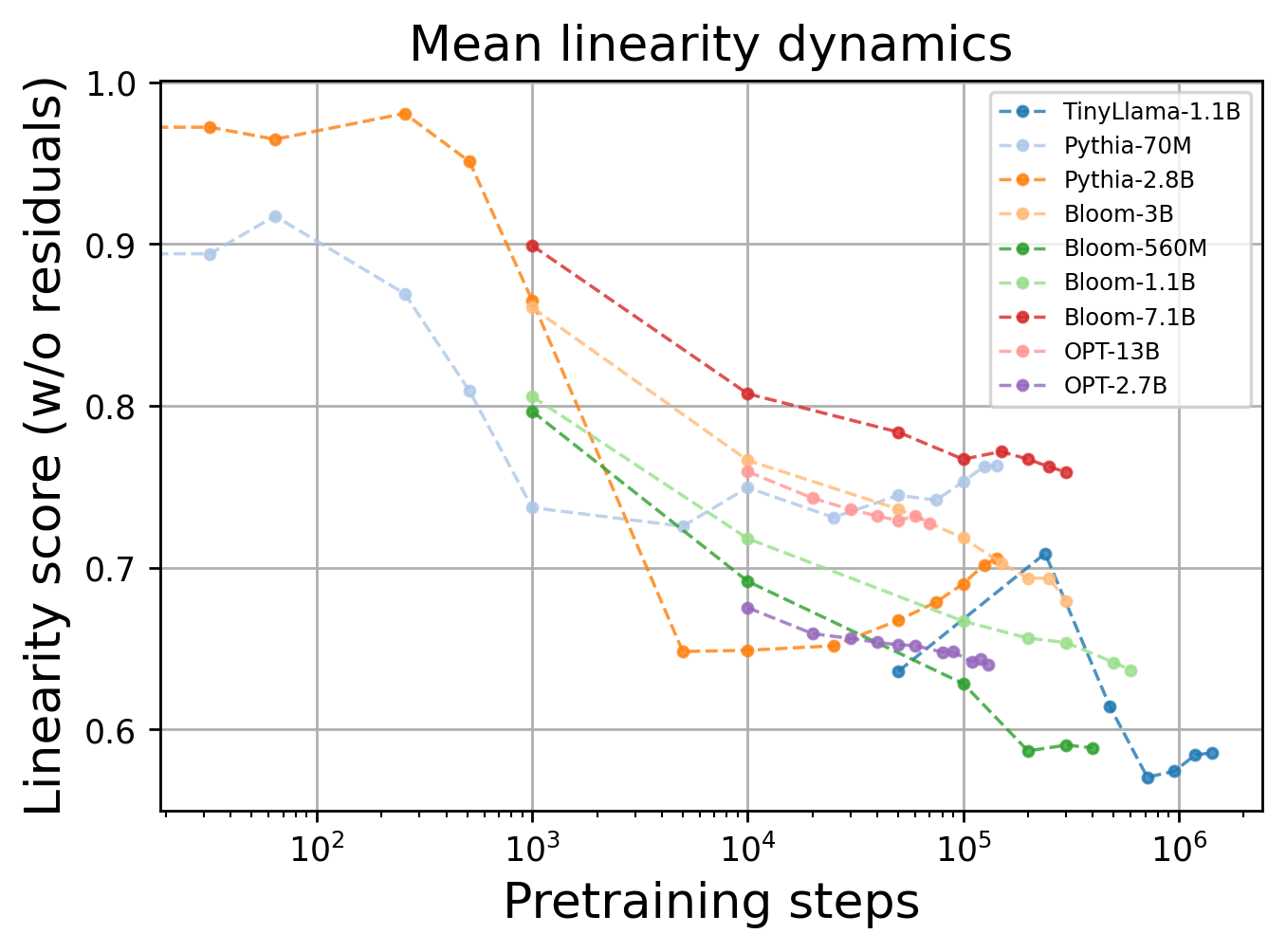}
    \caption{Linearity score (averaged across layers) at different pretraining steps of open source models.}
    \label{fig:mean_linearity}
\end{figure}

\begin{figure*}
    \centering
    \includegraphics[width=.9\linewidth]{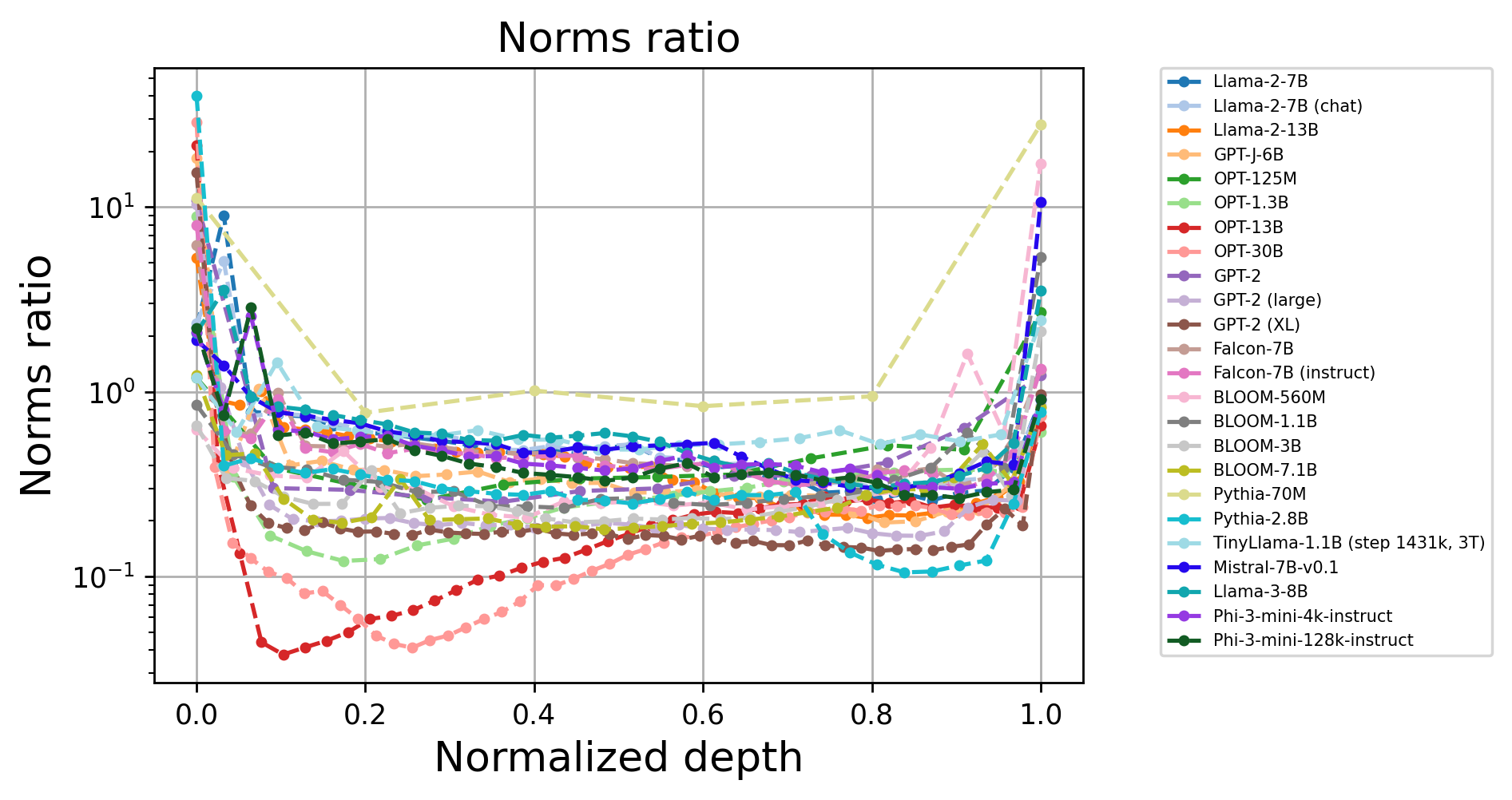}
    \caption{The relationship between transformer block output norm and resulted residual stream embedding norm.}
    \label{fig:norms}
\end{figure*}

In our study of the embedding properties of various layers of transformer decoders, we focus on understanding the degree of linearity and smoothness of transformations between sequential layers. 
\subsection{Linearity Score}
To determine the degree of linear dependence of two sets of vectors, we used a metric obtained by generalizing the Procrustes similarity \cite{ProcrustesGower} to the case of arbitrary linear transformations.

Let $X, Y \in \mathbb{R}^{n \times d}$ represent the centered sets of embeddings, to calculate linearity score we use normalized matrices $\tilde{X} = X / ||X||_2$, $\tilde{Y} = Y / ||Y||_2$ (where $||\cdot||_2$ denotes the Frobenius norm of the matrix) and defined

 \[
 \textrm{linearity\_score} := 1 - \min_{A \in R^{d \times d}}||\tilde{X}A - \tilde{Y}||_2^2
 \]

This is almost the same formula as in Procrustes similarity, the only difference is that, instead of considering the minimum among orthogonal transformations, we use the minimum among all linear transformations to find the optimal mapping in terms of squared errors.

We chose such approach for its robustness in evaluating the linearity of embeddings, especially considering the scale variance across transformer layers. Unlike $L_2$ norm, which lacks scale invariance, Procrustes normalization offers a bounded metric in the range [0,1].

Surprisingly, the linearity scores of layers in all tested transformer decoders were found to be close to 1, indicating a high degree of linearity in embedding transformations (Figure~\ref{fig:linearity_profiles}). 

This phenomenon can be partly explained by the observation that the norm of each block's contribution to the residual stream is remarkably low (Figure~\ref{fig:norms}). Moreover, when assessing the linearity of the main stream (embeddings w/o residual component) by subtracting the embedding values of each layer from the previous layer, one can notice that the degree of linearity significantly decreases (Figure~\ref{fig:linearity_profiles}). This suggests that the inherent linearity is not as straightforward as it is initially estimated. Moreover, the low norm contribution of individual blocks resulted in embeddings from adjacent layers being closely aligned in terms of cosine similarity.

One more insight is that the combination of seemingly linear blocks can lead to non-linear outcomes. \citet{superposition} suggests that complex features can be encoded across components of neural networks, applicable to attention heads in transformers. This indicates that the cumulative effect of linear transformations might enable the encoding of intricate non-linear representations.

Furthermore, our feature triggering regime hypothesis proposes that rare specific features on a few tokens with high non-linearity significantly influence model behavior — in the Figure \ref{fig:distribution} one can see that some layers of OPT-1.3B have the long tailed distribution of $L_2$ errors, which means that there are still sparse spikes of non-linearity. 

\citet{borse2023dejavu} explored how a sparse subset of model parameters can be dynamically activated for efficient inference, supporting the idea that within predominantly linear architectures, certain non-linear interactions are crucial for model functionality.

\subsection{Linearity Dynamics at Pretraining and Fine-tuning}

Our exploration extends to examining the linearity dynamics of both open-source models with publicly available intermediate checkpoints and our custom models trained on small datasets. Through this analysis, we aimed to understand the dynamics of linearity, especially in the main stream (contextualized embeddings including the residual component), across different stages of model training.

As illustrated in the Figure~\ref{fig:mean_linearity}, the analysis reveals a notable trend: as the models undergo pretraining, the linearity of the main stream gradually decreases on average. This phenomenon is consistently observed in all models examined, indicating a fundamental aspect of transformer-decoder learning dynamics.

In our analysis of the fine-tuning phase across diverse tasks, including those in the SuperGLUE benchmark \cite{wang2019superglue} and the reward-modeling task on the Anthropic-Helpful dataset \cite{bai2022training}, we notice an interesting change. Contrary to the decreasing trend of linearity observed during the pretraining phase, all models under study show an increase in linearity during fine-tuning. This finding indicates that task-specific fine-tuning tends to reinforce and amplify the linear characteristics of transformer models, as shown in Table~\ref{table:finetuning_linearity}.

\begin{table*}
\small
\centering
\begin{tabular}{|l|c|c|c|c|}
\hline
\textbf{Model Name} & \textbf{Super\_Glue/MultiRC} & \textbf{Super\_Glue/BoolQ} & \textbf{Super\_Glue/CB} & \textbf{Reward Modeling} \\ \hline
OPT-125M & \gradecolor{0.085} $\pm$ 0.008 & \gradecolor{0.217} $\pm$ 0.038 & \gradecolor{0.048} $\pm$ 0.009 & \gradecolor{0.060} $\pm$ 0.008 \\
OPT-1.3B & \gradecolor{0.055} $\pm$ 0.021 & \gradecolor{0.382} $\pm$ 0.004 & \gradecolor{0.088} $\pm$ 0.010 & \gradecolor{0.062} $\pm$ 0.007 \\
OPT-2.7B & \gradecolor{0.061} $\pm$ 0.025 & \gradecolor{0.356} $\pm$ 0.005 & \gradecolor{0.066} $\pm$ 0.029 & \gradecolor{0.054} $\pm$ 0.003 \\
Llama2-7B & \gradecolor{0.141} $\pm$ 0.006 & \gradecolor{0.051} $\pm$ 0.024 & \gradecolor{0.081} $\pm$ 0.070 & \gradecolor{0.194} $\pm$ 0.027 \\
GPT2 & \gradecolor{0.085} $\pm$ 0.021 & \gradecolor{0.048} $\pm$ 0.016 & \gradecolor{0.004} $\pm$ 0.003 & \gradecolor{0.092} $\pm$ 0.013 \\
GPT2-Large & \gradecolor{0.049} $\pm$ 0.003 & \gradecolor{0.023} $\pm$ 0.008 & \gradecolor{0.025} $\pm$ 0.014 & \gradecolor{0.085} $\pm$ 0.008 \\
GPT2-XL & \gradecolor{0.040} $\pm$ 0.007 & \gradecolor{0.037} $\pm$ 0.007 & \gradecolor{0.028} $\pm$ 0.019 & \gradecolor{0.038} $\pm$ 0.008 \\ \hline
\end{tabular}
\caption{Delta of linearity score w/o residuals after fine-tuning various tasks. Note that all values are strictly positive, which means that linearity always increases during fine-tuning.}
\label{table:finetuning_linearity}
\end{table*}

In fine-tuning, we train models on three NLI tasks from the SuperGLUE benchmark: MultiRC, BoolQ, and CB, treating them as binary text classification challenges. In the BoolQ task, for instance, we combine the question and the passage into a single text, marking them with "question:" and "passage:" respectively, and consider the binary answer as the classification label.

Reward models trained on text pairs with contrastive loss \cite{ouyang2022training} demonstrate a similar trend in linearity scores, proving even more stability across different seed values.

\begin{table*}[ht]
\tiny
\centering
\begin{tabular}{|l|c|c|c|c|c|c|c|c|c|c|c|}
\hline
\textbf{Model/Task} & \textbf{boolq} & \textbf{cb-acc} & \textbf{cb-f1} & \textbf{copa} & \textbf{multirc} & \textbf{record-f1} & \textbf{record-em} & \textbf{rte} & \textbf{wic} & \textbf{xstorycloze-en} & \textbf{mean} \\ \hline
Mistral 650M & 48.50 & \textbf{42.86} & 21.96 & 56.0 & 56.97 & 21.80 & 21.05 & 51.26 & \textbf{51.10} & 61.75 & 43.33 \\
Mistral 650M + cosine (0.5) & \textbf{57.50} & 41.07 & \textbf{28.57} & \textbf{61.0} & \textbf{57.10} & \textbf{23.20} & \textbf{22.54} & \textbf{55.23} & 50.00 & \textbf{64.39} & \textbf{46.06} \\ \hline\hline
Mistral 150M & 38.84 & \textbf{42.86} & \textbf{27.39} & 56.0 & 44.16 & 20.07 & 19.42 & 51.26 & 51.10 & 59.89 & 41.10 \\
Mistral 150M + MSE (0.5) & 38.84 & 39.29 & 19.30 & 60.0 & \textbf{57.59} & 20.46 & 19.77 & \textbf{53.07} & 50.47 & 57.64 & 41.64 \\
Mistral 150M + MSE (2.0) & 39.39 & 41.07 & 19.41 & 57.0 & 46.53 & \textbf{22.62} & \textbf{21.89} & 51.99 & 50.00 & 56.52 & 40.64 \\
Mistral 150M + cosine (0.5) & \textbf{44.16} & 37.50 & 24.18 & \textbf{62.0} & 54.54 & 21.67 & 20.99 & 50.90 & 50.47 & \textbf{61.35} & \textbf{42.78} \\ \hline
\end{tabular}
\caption{SuperGLUE results.}
\label{tab:superglue_mistral}
\end{table*}

\section{Improving Linearity with Regularized Pretraining}

Aiming to understand the impact of linearity on transformer models, we embark on pretraining experiments using the Mistral architecture with model sizes of 150M, and 650M. These models are pretrained on carefully selected clean datasets, TinyStories \cite{eldan2023tinystories} and Tiny-textbooks \cite{li2023textbooks}, chosen for their diverse and rich content, which has been proven to be suitable for fast training of the small models \cite{Zhao2023BabyStoriesCR} and architecture experiments \cite{Sharifnassab2024MetaOptimizeAF}.

We introduce specific loss terms to adjust the relations between embeddings within transformer layers:
\begin{itemize}

    \item \textbf{MSE regularization term:} Experimentation with mean squared error (MSE) loss between embeddings of consecutive layers, designed to minimize the distance between these embeddings, thereby promoting consistency across the layers.
        \[L_{MSE} = \lambda \sum (\|emb_{i} - emb_{i-1}\|^2).\]
    
    \item \textbf{Cosine Similarity regularization term:} The application of a cosine-based regularization that encourages contextualized embeddings from sequential layers to align closer to each other, effectively reducing their angular difference to zero on average.
\[L_{cosine} = \lambda \sum (1 - \cos(emb_{i}, emb_{i-1})).\]
    
\end{itemize}

\begin{figure}
    \centering
    \includegraphics[width=\linewidth]{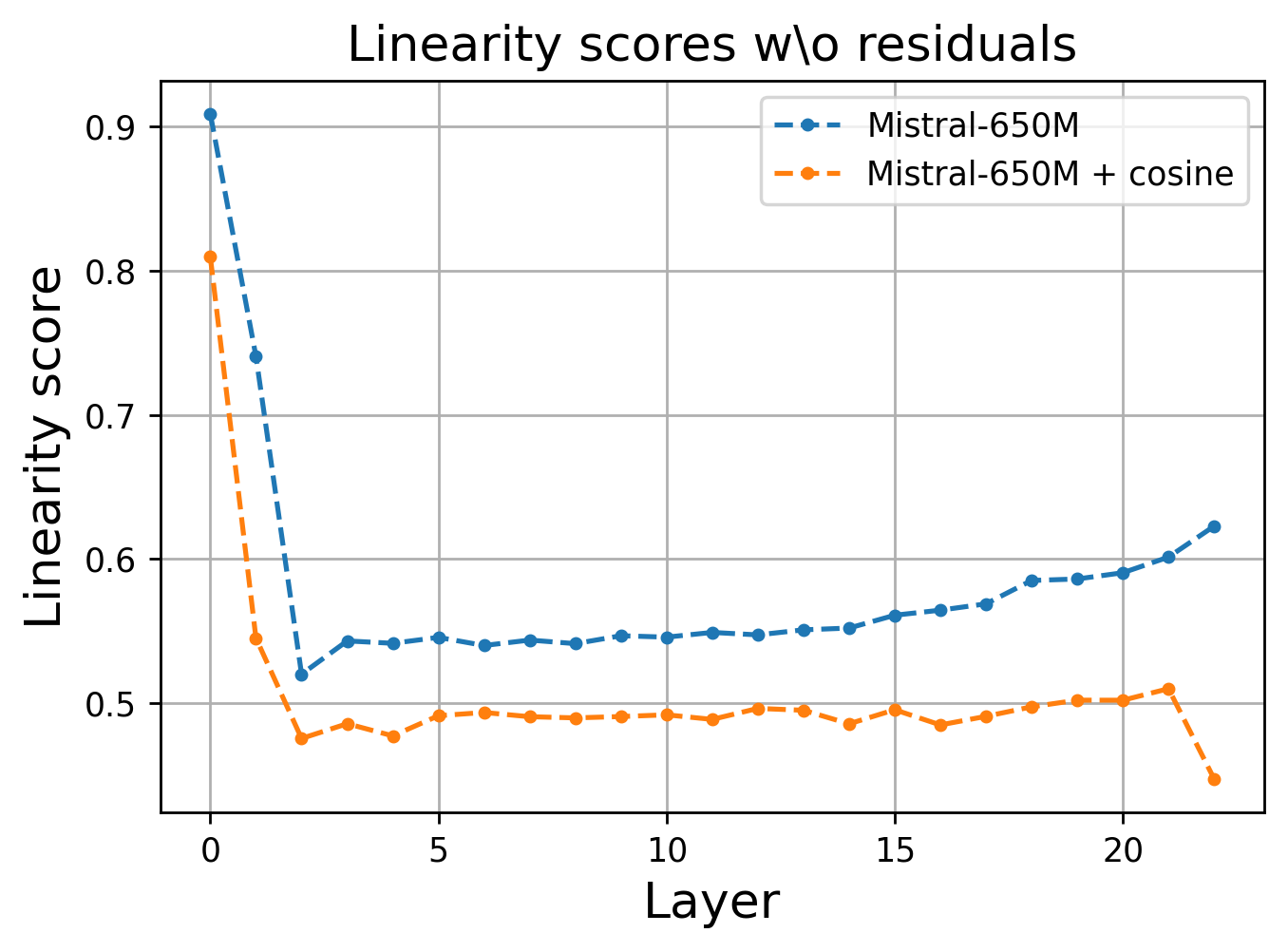}
    \caption{Linearity score of different layers with and without cosine regularization used at pretraining.}
    \label{fig:lin_probing}
\end{figure}

\begin{figure}
    \centering
    \includegraphics[width=\linewidth]{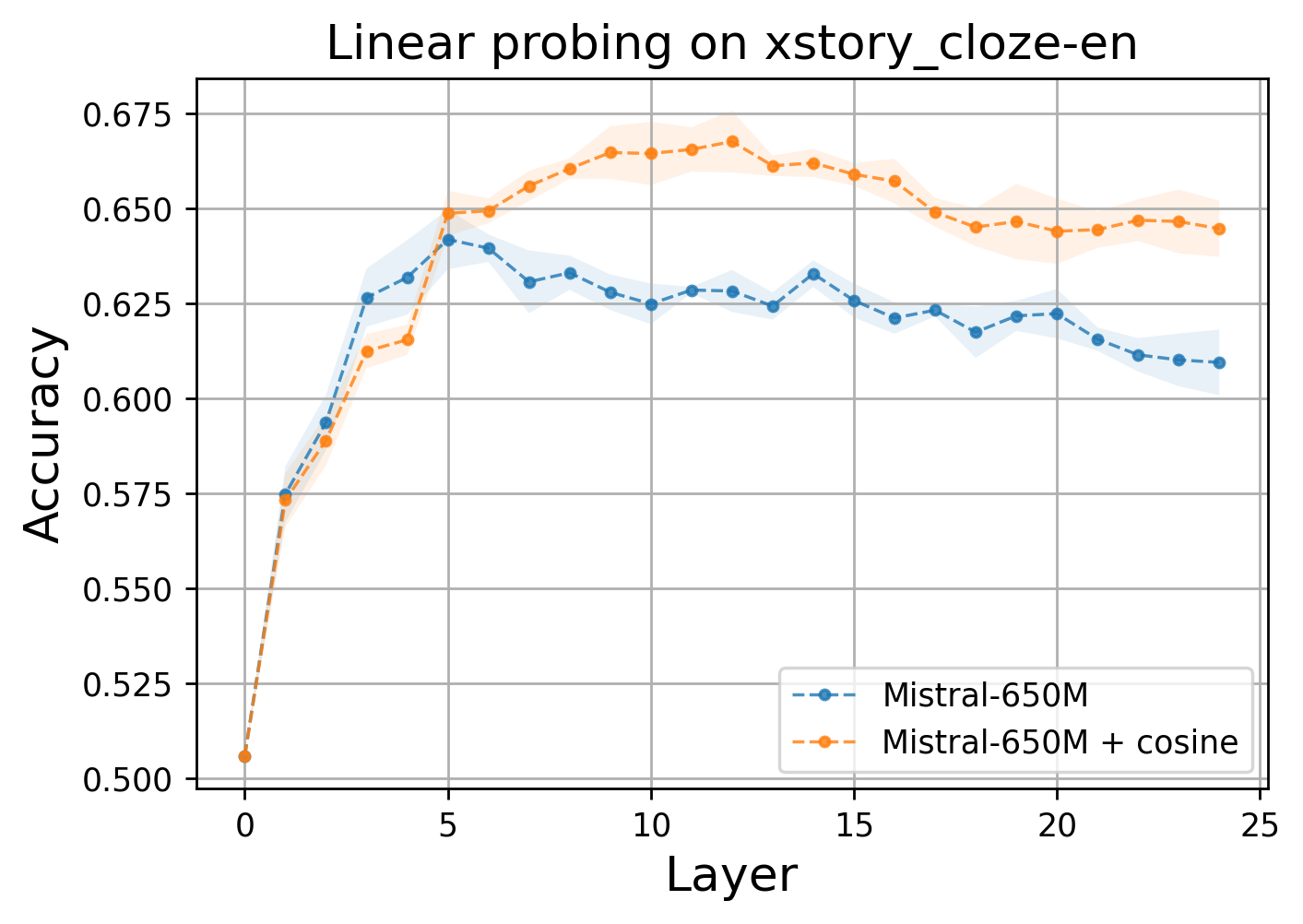}
    \caption{Linear probing of embeddings from different layers of Mistral-650M pretrained with and without suggested cosine regularization.}
    \label{fig:lin_reg}
\end{figure}

The most promising results are achieved using a cosine-based approach that encourages the embeddings of sequential layers to converge, effectively making the cosine similarity between them closer to 1 on average. This method shows significant perspectives in the enhancing model performance. We evaluate the effectiveness of our approach through validation using GPT-4 on TinyStories prompts according to the \citet{eldan2023tinystories}  methodology, linear probing techniques, and evaluation on SuperGLUE benchmarks. The results are presented in the Table~\ref{tab:superglue_mistral} and Table~\ref{table:tiny_stories}. As it can be seen in the Figure~\ref{fig:lin_reg}, linearity scores are lower at each layer of the model after pretraining with such regularization.

To further assess the expressiveness of embeddings across different layers, we conducted linear probing on outputs of all the layers of the Mistral-650M model, both pretrained with and without cosine regularization, on the xstorycloze-en task from SuperGLUE. The results clearly indicate that embeddings from the model pretrained with regularization exhibit better performance compared to those from the standard model (Figure~\ref{fig:lin_probing}).

This contradictory outcome, where the term appears to draw embeddings from neighbouring layers closer together, making them more similar in terms of cosine similarity, has prompted a deeper investigation. Our observations suggest that as embeddings become more similar across layers, the model may compensate for the reduction in variability by amplifying non-linear processing capabilities in the residual stream. Although this hypothesis requires further exploration, it offers a fascinating insight into the adaptive mechanisms of transformer models in response to altered internal dynamics.

\begin{table}
\tiny
\centering
\begin{tabular}{|l|c|c|c|c|c|}
\hline
\textbf{Mistral config} & \textbf{Grammar} & \textbf{Creativity} & \textbf{Consistency} & \textbf{Plot} & \textbf{Mean} \\
\hline
650M & 5.47 & 6.60 & 4.81 & 4.67 & 5.39 \\
650M + cosine (0.5) & \textbf{6.07} & \textbf{7.02} & \textbf{5.74} & \textbf{5.48} & \textbf{6.08} \\
\hline
\hline
150M & 4.88 & 6.51 & 4.16 & 3.88 & 4.86 \\
150M + MSE (0.5) & \textbf{5.19} & 6.70 & 4.47 & 4.20 & 5.14 \\
150M + MSE (2.0) & 5.00 & 6.81 & 4.56 & 4.29 & 5.17 \\
150M + cosine (0.5) & 5.14 & \textbf{6.91} & \textbf{4.77} & \textbf{4.95} & \textbf{5.44} \\
\hline
\end{tabular}
\caption{TinyStories prompts completions evaluation via GPT-4}
\label{table:tiny_stories}
\end{table}

\section{Exploiting Linearity for Pruning}

\begin{figure}
    \centering
    \includegraphics[width=\linewidth]{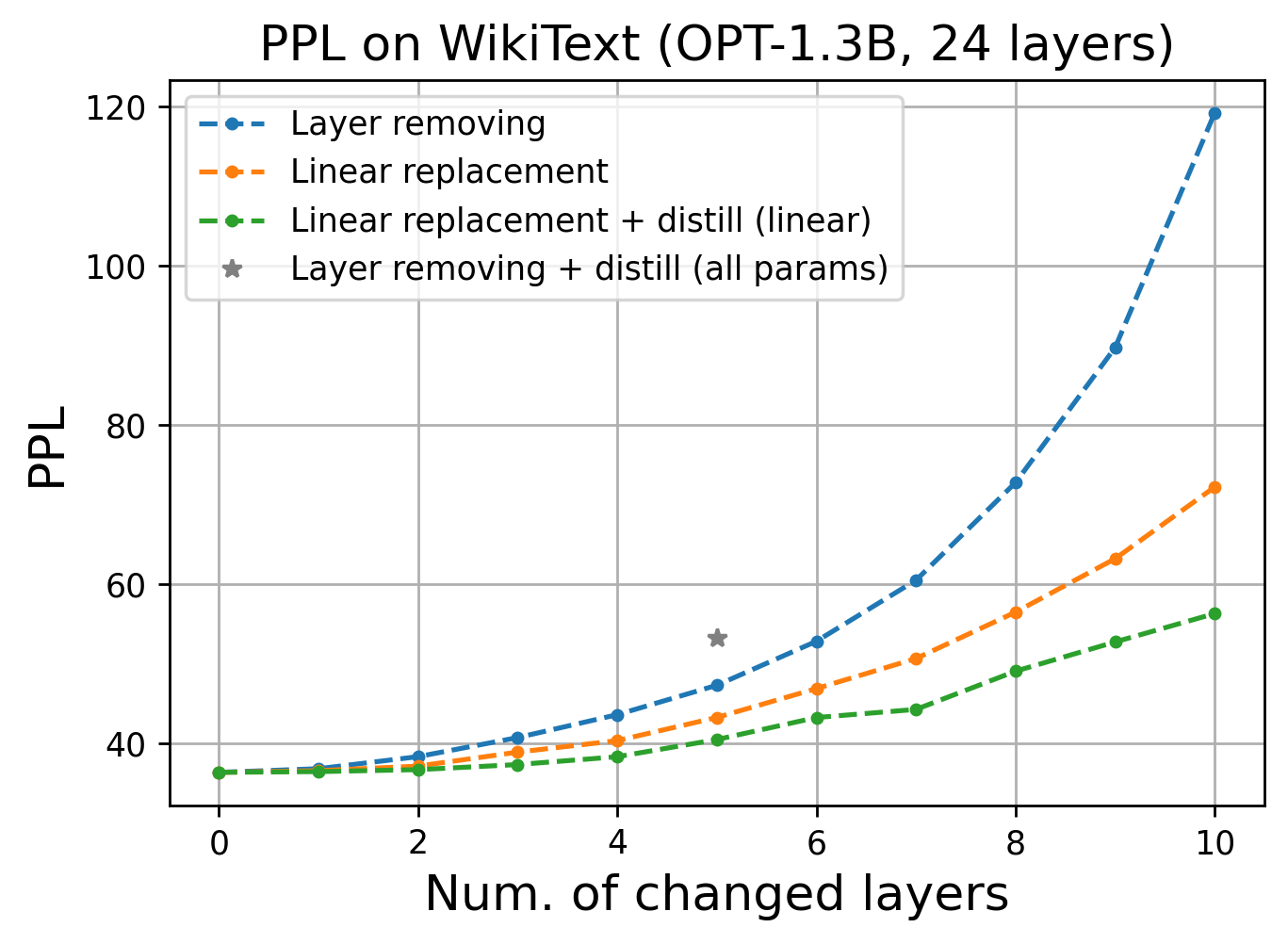}
    \caption{Perplexity on WikiText for various pruning and distillation methods (lower is better).}
    \label{fig:ppl}
\end{figure}

\begin{figure}
    \centering
    \includegraphics[width=\linewidth]{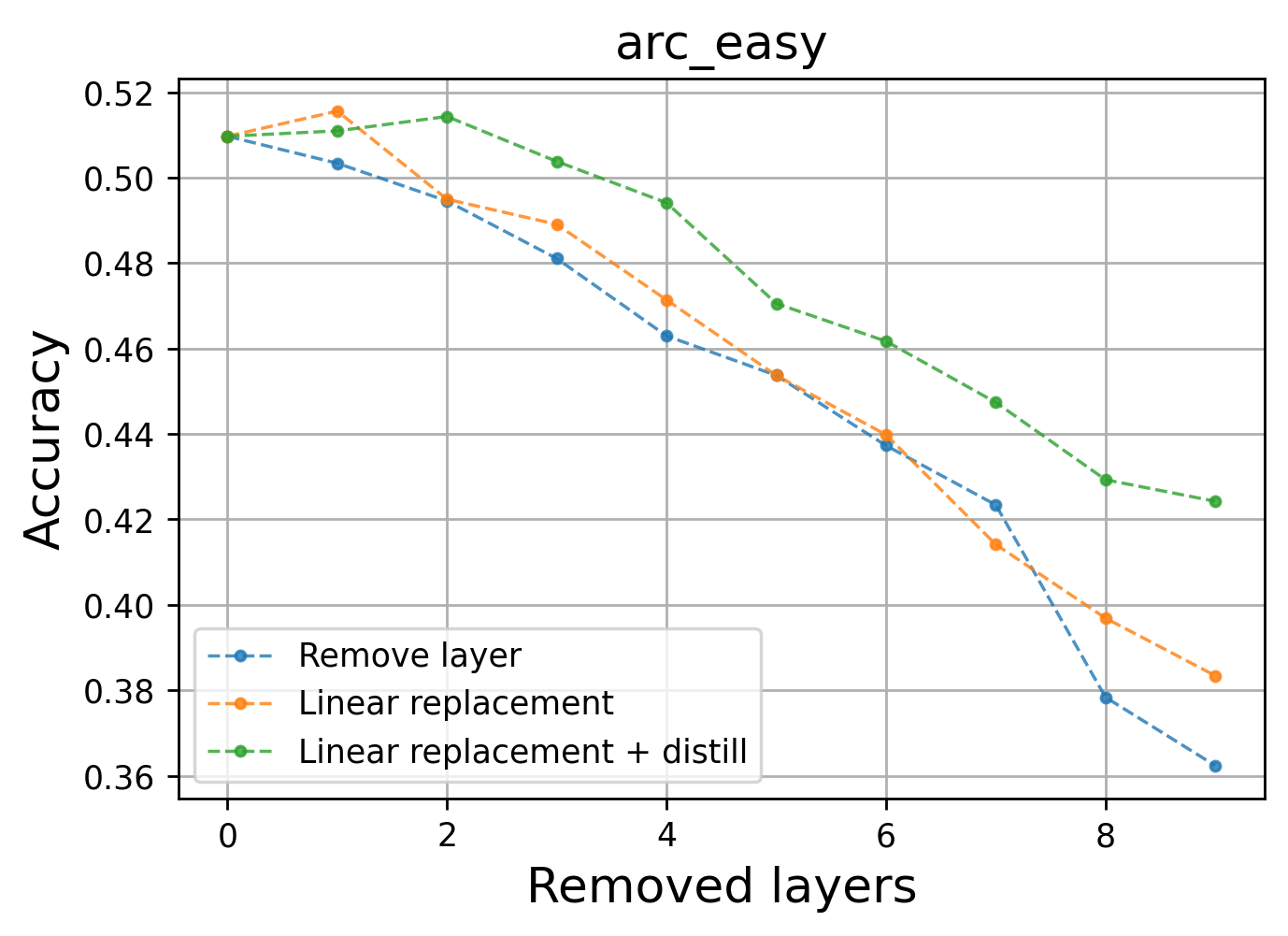}
    \caption{OPT-1.3B results on ARC-easy dataset with suggested pruning techniques.}
    \label{fig:pruning}
\end{figure}

Leveraging the inherent linearity of transformer layers, we explore a pruning strategy that sequentially removes the most linear layers. This approach allows you to reduce the size of the model slightly by removing just a few layers without significantly compromising performance. Further enhancement of this strategy involves replacing the pruned layers with linear approximation and incorporating a distillation loss (specifically MSE layerwise) to minimize performance degradation. The training focuses on these linear replacements, fine-tuning them to effectively mimic the original layers' function. The effectiveness and the impact of these methods are detailed in the Figure~\ref{fig:pruning}. We use TinyStories for linear approximation and distillation training stage. As it can be seen in the Figure~\ref{fig:ppl}, perplexity is less affected by pruning with linear replacements and following distillation compared to just removing transformer layers.

\begin{figure}
    \centering
    \includegraphics[width=\linewidth]{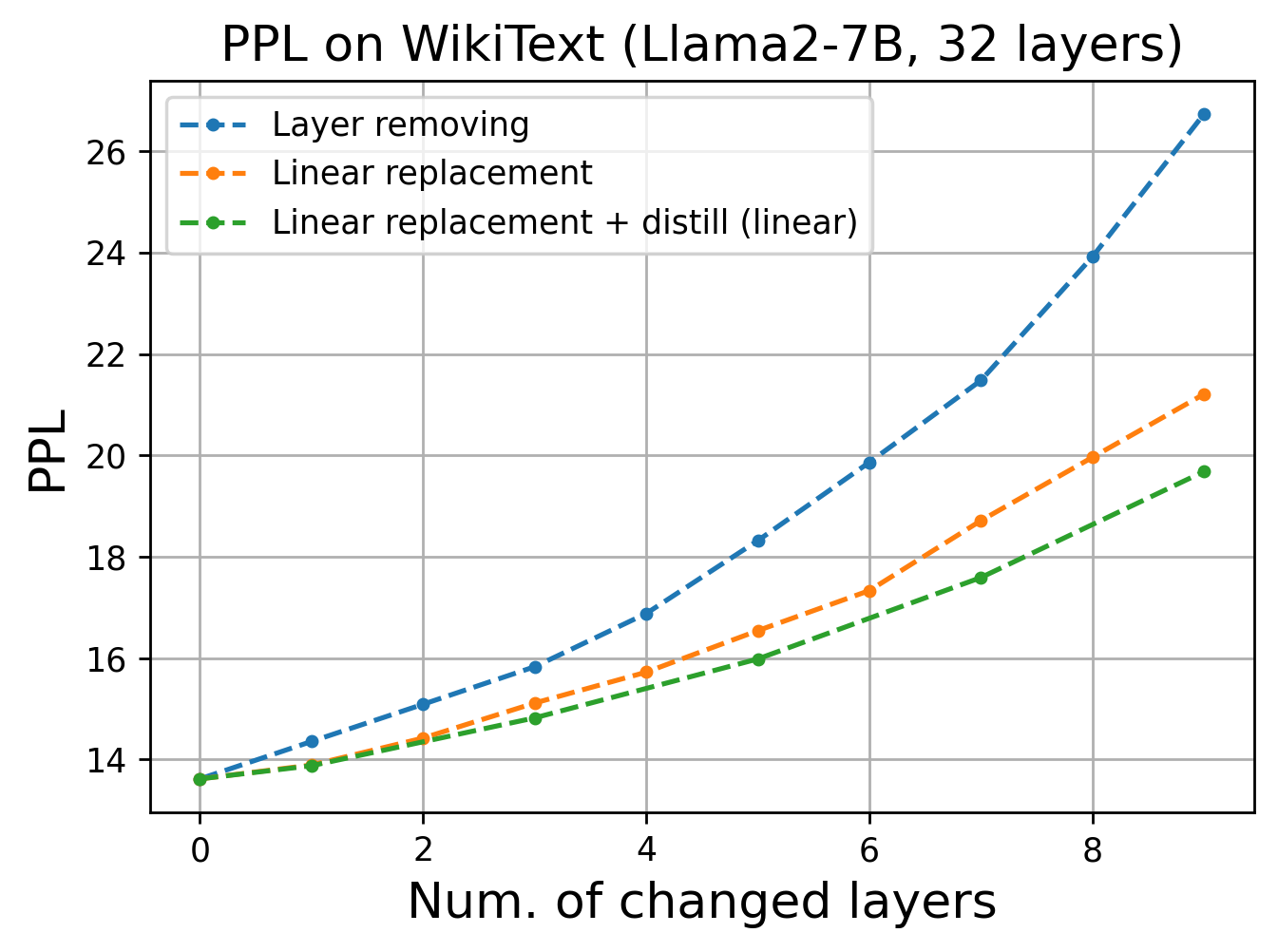}
    \caption{Perplexity on WikiText for various pruning and distillation methods (lower is better).}
    \label{fig:llama_prune}
\end{figure}

\section{Conclusion}
In our study we provide an in-depth exploration of linearity within transformer decoders, revealing their inherent near-linear behavior in various models. We discover that while pretraining tends to increase nonlinearity within layers, fine-tuning on specific tasks can paradoxically reduce it. 
We propose new pruning and distillation techniques inspired by previous observations, demonstrating that it is possible to refine and optimize transformer models without compromising their performance. 
The suggested cosine-based regularization approach during pretraining further contributes to model efficiency and performance on benchmarks such as SuperGLUE and TinyStories, while reducing the linearity of its layers (w/o residual components). 

Our study highlights the significant relationship between linearity and performance of transformer decoders, offering strategic guidance for future developments in the efficiency and flexibility of these models.

\section{Limitations}

Despite the promising advancements presented in this study, it is essential to acknowledge its limitations. Firstly, our analysis predominantly focuses on transformer decoders, thus the generalizability of our findings to encoder-only or encoder-decoder architectures may be limited.

Secondly, the depth pruning and distillation techniques, while being effective in our experiments, were evaluated within a specific set of conditions and models. The scalability of these methods to larger, more complex models or different domains is yet to be fully ascertained. 

Moreover, the new regularization approach aimed at pretraining demonstrates potential, yet its effectiveness across a broader spectrum of tasks requires further validation.

\section{Ethics Statement}

We are committed to ethical principles for AI research, focusing on transparency and responsible experimentation. Our research, while suggesting efficiency improvements, prompts consideration of implications such as privacy and fairness.


\bibliography{custom}

\clearpage
\appendix
\section{Error Distribution by Layers}
\label{sec:appendix}

\begin{figure*}[ht]
    \centering
    \includegraphics[width=\textwidth]{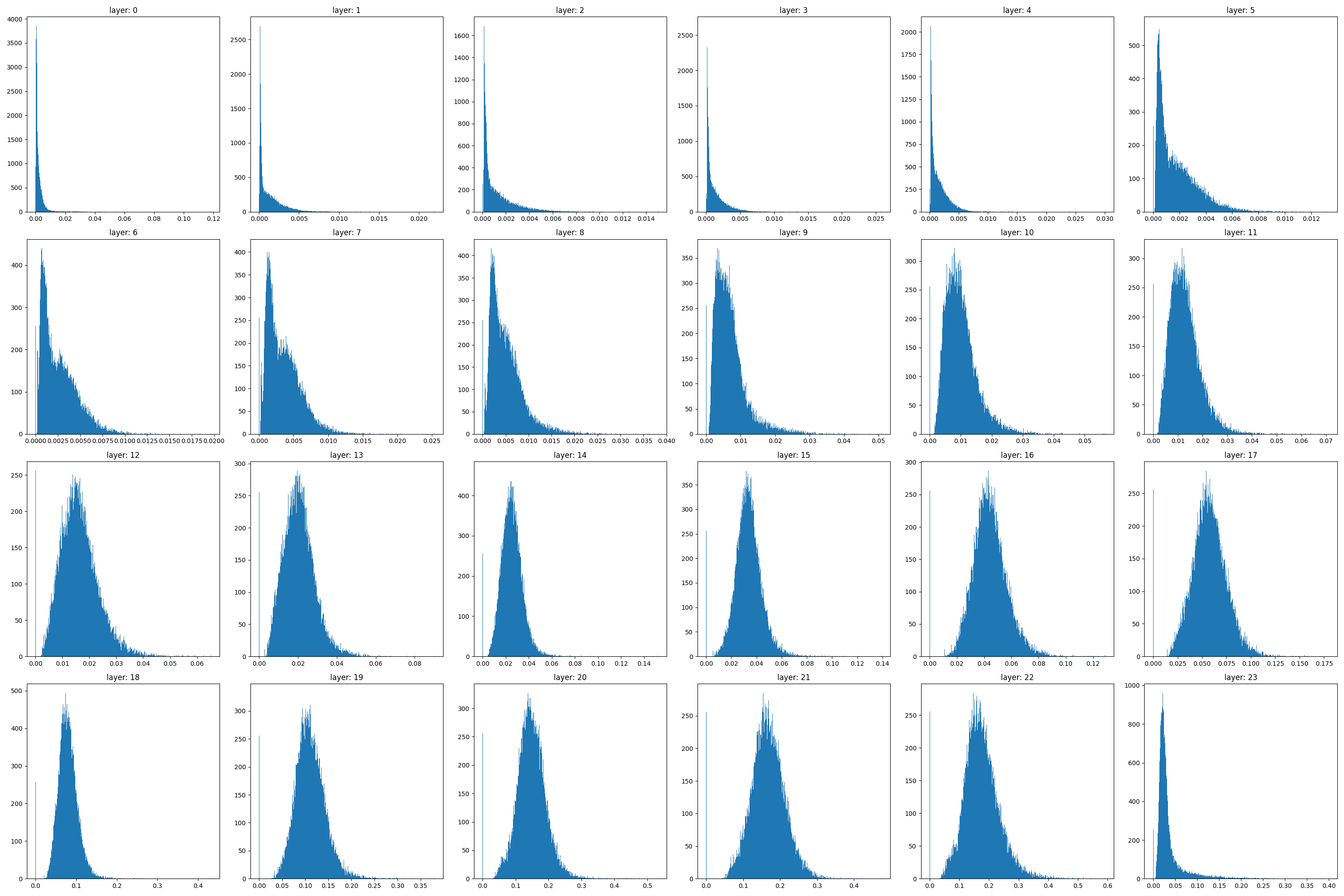}
    \caption{$L_2$ error distribution of linear approximation across different layers of OPT-1.3B.}
    \label{fig:distribution}
\end{figure*}

In the Figure \ref{fig:distribution} we present a visualization of $L_2$ error distribution across several layers of OPT-1.3B decoder architecture.

\end{document}